\documentclass[conference]{IEEEtran}
\usepackage{blindtext, graphicx}
\usepackage{multicol}
\usepackage{color}
\usepackage{url}

\usepackage[backend=bibtex,style=ieee,sorting=ynt]{biblatex}
\begin{document}
\title{End-to-End Radio Traffic Sequence Recognition with Deep Recurrent Neural Networks}

\author{\IEEEauthorblockN{Timothy J. O'Shea}
\IEEEauthorblockA{Bradley Department of Electrical \\ 
and Computer Engineering\\
Virginia Tech, Arlington, VA\\
Email: oshea@vt.edu}
\and
\IEEEauthorblockN{Seth Hitefield}
\IEEEauthorblockA{Bradley Department of Electrical \\
and Computer Engineering\\
Virginia Tech, Arlington, VA\\
Email: hitefield@vt.edu}
\and
\IEEEauthorblockN{Johnathan Corgan}
\IEEEauthorblockA{Corgan Labs, \\
San Jose, CA\\
Email: johnathan@corganlabs.com}
}

\maketitle

\begin{abstract}
We investigate sequence machine learning techniques on raw radio signal time-series data.  By applying deep recurrent neural networks we learn to discriminate between several application layer traffic types on top of a constant envelope modulation without using an expert demodulation algorithm.  We show that complex protocol sequences can be learned and used for both classification and generation tasks using this approach.
\end{abstract}


\IEEEpeerreviewmaketitle

\section{Introduction}

Traffic analysis and deep packet inspection are important tools in ensuring quality of service (QoS), network security, and proper billing and routing within wired and wireless networks.  Systems and algorithms exist today to discern between different protocols and applications for these reason, but new methods provide great potental for improvement.

Current day techniques often involve the use of numerous brittle protocol parsers which must parse a combinatorially large number of different network and application protocols, limiting parsing abilities to known protocols whose parsers have been manually implemented, potentially with parser implementation vulnerabilities or other defects.  On top of protocol parsing, wireless signals also require detection, synchronization, equalization, symbol to bit de-mapping and error correction decoding.  Each of these algorithms adds complexity, implementation cost, vulnerability potential, and protocol specificity to the ultimate solution under development.

By applying machine learning to the task of interpreting modulated radio signals carrying high level protocols directly, we demonstrate that we can successfully treat this demapping and interpretation process as a learned data mapping process within a machine learning framework.  In doing so we form a model which can learn to generalize and to make decisions on new unseen modulations and protocols.  We build a model which is not prone to trivial parser based security vulnerabilities and we form a model which does not incur cost and complexity to development which scales with the number of specific protocols implemented since they are derived from datasets using a model that generalizes well.  We have previously demonstrated \cite{convmodrec} that this class of approach using deep neural networks to learn a radio discrimination task on low level modulations can be highly effective, but in this work we show that this potential also spans up the stack to higher layer traffic types as well.

\subsection{Recurrent Networks in Natural Language}

\begin{figure}[h]
\label{fig:spotify}
\caption{Basic LSTM Unit Transfer Function Diagram from \cite{lstm_diag}}
\centering
\includegraphics[width=0.3\textwidth]{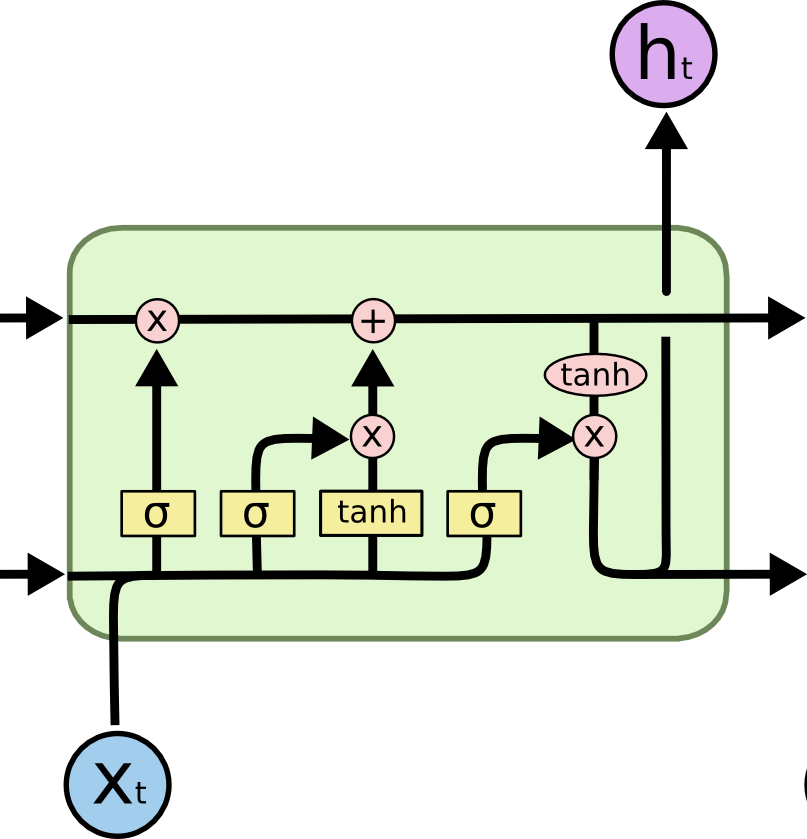}
\label{fig:lstm1}
\end{figure}

Recurrent neural network approaches to temporal sequence learning are not a new thing, they have been very successful in recent years in natural language translation, natural language embedding tasks for information retrieval or mapping, and automatic voice recognition fields among other applications.  In each of these, sequences of tokens, either characters or phonemes are encoded using recurrent neural networks such as the long short-term memory \cite{lstm} (LSTM).  Recurrent neural networks based on the simple recurrent unit, the LSTM, and the Gated Recurrent Unit (GRU) \cite{gru} are all widely used and their capacity for sequence learning is quite impressive as is visable in a task as simple as presenting natural language text characters to such a system \cite{charrnn}.  The LSTM basic neuron unit's transfer function and structure is shown in figure \ref{fig:lstm1}.

Many applications have also successfully employed recurrent networks for translation between sequence domains (such as different languages) \cite{sequencetosequence} based on embeddings, mapping from sequences to discrete classes \cite{sequencetoclass}, and many other sequence related tasks.  The LSTM has been especially widely used in this field, as a highly successful recurrent network primitive, but does not represent the only or the least computationally expensive choice as the simple RNN and GRU are both used widely.  In both voice and text modeling fields, state of the art methods which used to leverage Hidden Markov models (HMMs) for sequence prediction have been largely replaced with this class of RNN based approach to modeling.

\subsection{Background on Radio Sequence Motivations}

In radio communications, the radio transmitter and receiver are comprised of a number of sequence to sequence translation routines \cite{ted}.  These translate between sequences of protocol data bits, forward error corrected encoded bits, randomized and whitened bits, framed bits, and finally to modulated and encoded symbols which directly traverse the radio channel.  

Rather than implementing expert algorithms for each of these, we can attempt to learn these sequence translation mappings by presenting data to an appropriate machine learning architecture.  Ideally learning to consume radio symbols, process idle-traffic patterns, data framing patterns, and data payload patterns all directly from the example data sequences presented to the learning algorithms, rather than relying on any amount of expert encoding and decoding algorithm descriptions.

\section{Supervised Traffic Type Learning}

In our network we train a multi-layer LSTM-based sequence learner network on a succession of slices of our modulated radio signal to perform supervised classification into one of 11 different protocol traffic classes. 

We an architecture where LSTM units operate directly on complex base-band I/Q signal representations where I and Q components are treated as seperate and independent channels, followed by fully connected layers using linear rectifiers and softmax activation on the final output layer.  

\subsection{Dataset Generation}

We generate a data set comprising several different common network application protocols transmitted over a wireless link.  We first capture network traffic corresponding to the network activity behaviour of interest.  The applications selected are shown in the table below, including traffic from multimedia streaming, typical browsing and file downloading, software development, and system administration tasks.

\begin{itemize}
    \item {\bf Streaming}
    \begin{itemize}
        \item Video Streamin (via ABC video)
        \item Video Streaming (via Youtube)
        \item Music Streaming (via Spotify)
    \end{itemize}
    \item {\bf Utilities}
    \begin{itemize}
        \item Apt-get
        \item ICMP Response Test (Ping)
        \item Version Control (git)
        \item Internet Relay Chat (IRC)
    \end{itemize}
    \item {\bf Downloading/Browsing}
    \begin{itemize}
        \item Bit-Torrent
        \item Web browsing
        \item File transfar protocol (FTP)
        \item HTTP Download
    \end{itemize}
\end{itemize}

\emph{Wireshark} and \emph{tcpdump} were used to capture network traffic and generate traces of each network protocol used later in for training and classification.  While these utilities can be used to target specific network traffic (i.e. recording a specific port/connection/protocol), a more general capture provides additional behavioral data that would be useful for training and recognition, i.e. background traffic exists and related traffic such as domain name look-ups are occurring as well.  This provides a more realistic picture of what complex heterogeneous network traffic looks like rather than a setup which may have explicitly tried to capture isolated network traffic using just a single protocol.  It is also a challenge because the traffic of interest is not occurring at all times within the dataset, leading to some time windows which contain no information about the classification task of interest.

\begin{figure}[h]
\label{fig:capture}
\caption{Packet Capture Setup with an Isolated Virtual Machine}
\centering
\includegraphics[width=0.35\textwidth]{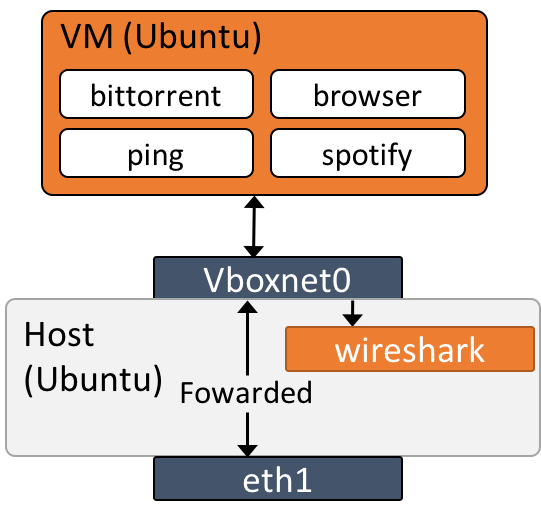}
\end{figure}

The setup we used for capturing network traffic is shown in Figure \ref{fig:capture}. Our goal was to isolate traffic while performing each task on a virtual machine to the network traffic originating on the host.  By connecting the virtual machine to a host-only network and enabling forwarding, all traffic over the interface can be easily captured with Wireshark or Tcpdump on the host system. Network address translation (NAT) is used to allow the guest system to access the Internet.  An example capture of a music streaming service (Spotify) can be seen in Figured \ref{fig:spotify}.

\begin{figure}[h]
\label{fig:spotify}
\caption{Wireshark Packet Capture of a Spotify Session}
\centering
\includegraphics[width=0.45\textwidth]{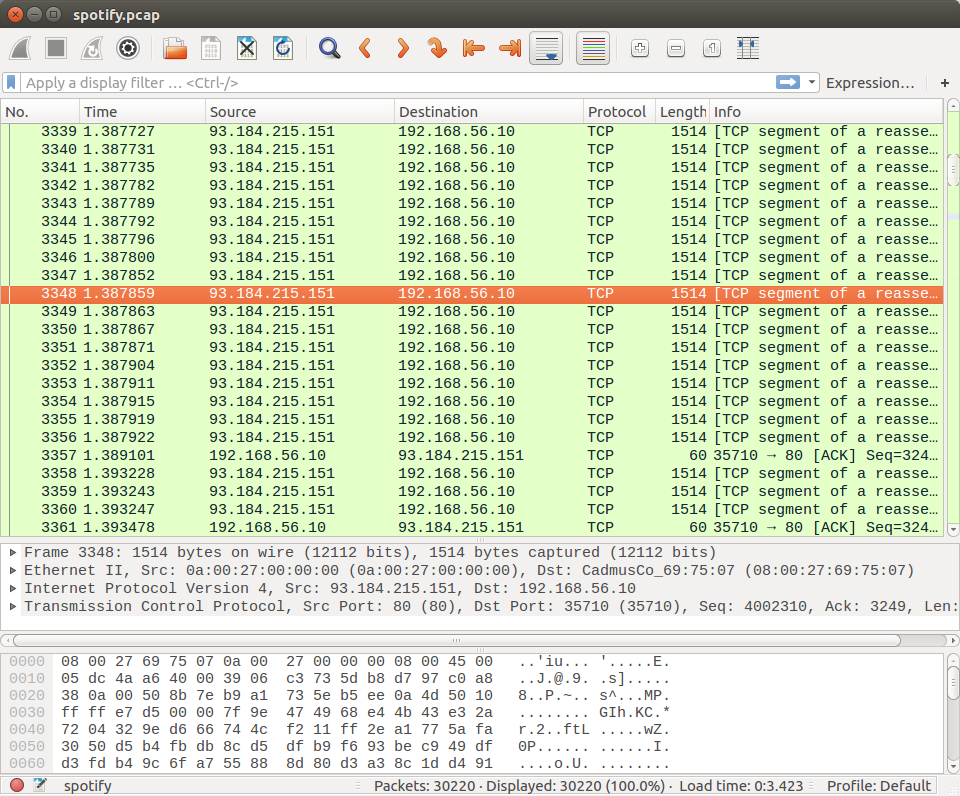}
\end{figure}

\begin{figure}[h]
\label{fig:mod}
\caption{Packet Capture Transmitter Flowgraph in GNU Radio}
\centering
\includegraphics[width=0.5\textwidth]{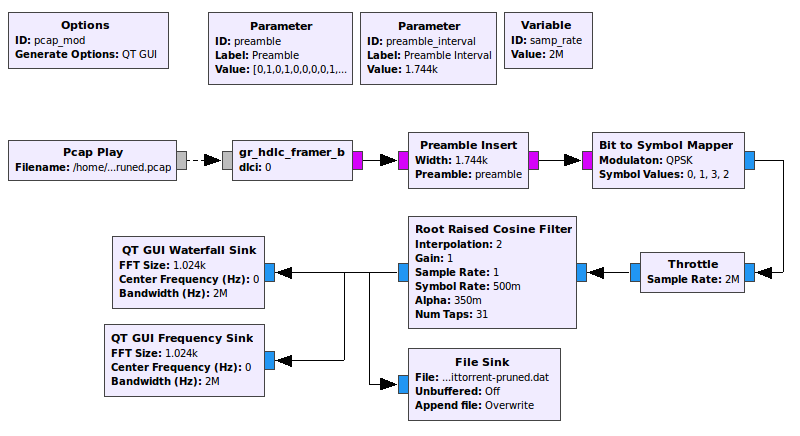}
\end{figure}

Once the network traffic was captured, the next step for generating the data sets was replaying the traffic through a continuously modulation modem and recording the IQ sample data to form our dataset.  The transmitter we use is a GNU Radio \cite{gnuradio} flow-graph that uses High-Level Data Link Control (HDLC) for framing and a Quadrature Phase Shift Keying (QPSK) for modulation without any error correction or randomization (shown in Figure \ref{fig:mod}.
The gr-pcap out-of-tree module (OOT) was used to replay packet captures with appropriate timing information in tact for each packet.  Messages are framed into the constant rate HDLC bit-stream by an HDLC framer, which constantly transmits the idle flag 0x7E if no input data is available.  This makes the classification task interesting because something is always being transmitted, a classifier can not simply learn the power envelope to identify protocol timing as is possible in a bursty CSMA/CD system.  A preamble is inserted periodically every 1744 bits to allow for PHY synchronization by a receiver and a throttle block is used to impose the desired baud rate.  By selecting different baud rates using this throttle, the constant data rate in the PCAP file varies from high or low percent utilization on the link, effecting the mix between idle and non-idle traffic.  We select a bit rate of around 1MBit/s which provides a reasonable middle ground on link utilization averaged over all of the different protocol recordings.  Bits are then mapped to QPSK symbols, passed through a root-raised-cosine filter and then "transmitted".  Here we simply save IQ symbols to a data file to be used in training and test. 

\subsection{Model Data Ingest}

For training models on this large time-series, we must chose how to present the data to the RNN model.  There are two considerations here, first how to slice a sequence into time steps to present to the sequence model and second how to partition the data on a macro scale into regions of training and test data.

In the first case, consider a time series $x(n)$ where we wish to create examples from linear subsequences.  In this case, we extract N windows of size L at a stride of M to form a three dimensional example vector.  In this case, the dimensions are expressed in the form of a real-valued tensor of shape Nx2xL, where the first dimensions is over window, the second is over the I/Q dimension, and the third is over time within each window.  Each tensor example is then formed from $L+(N-1)*M$ complex samples in the original time-series.  Since an optimal slicing is not known offhand for either task, we will use this notation throughout to refer to our input tensor shape tested during training.  We perform this slicing using python-numpy and ingest tensor data into Keras \cite{keras} and Theano \cite{theano} for model training.

For our supervised network-task classification model, we use one-hot target labels for each example where 1xK output values are all zero except where the target index k is of the example class, where it is set to 1.0.  This is commonly used along with a SoftMAX output activation layer to help in training for class prediction, and we use it the same way here.

In the case of a generative regression model, we use the same N time-step windows as out input tensor data, while using an N+1'th time-step of real-valued samples as our example target.

Lastly, as a pre-processing step, we consider whether to input I/Q samples, R/$\theta$ samples, R-only, or $\theta$ only from our sample representation, where R,$\theta$ represent the polar form of the I/Q sample.  We do this to consider capturing the circular relationship between in-phase and quadrature components which is thrown away when treating them as real valued separate channels. 

\begin{figure}[h]
\caption{1024 time samples of Spotify class}
\centering
\includegraphics[width=0.5\textwidth]{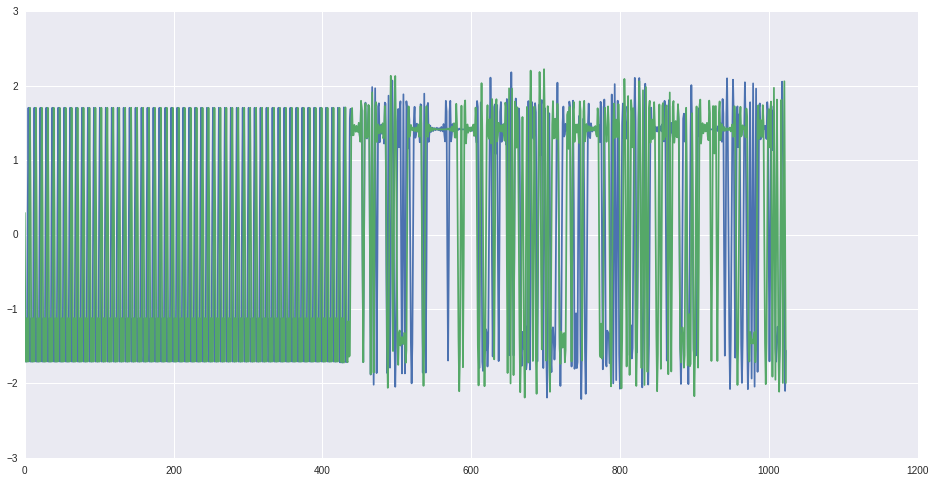}
\end{figure}

\subsection{Discriminative Model Training}

\begin{figure}[h]
\caption{LSTM256 Recurrent Network Structure}
\centering
\includegraphics[width=0.5\textwidth]{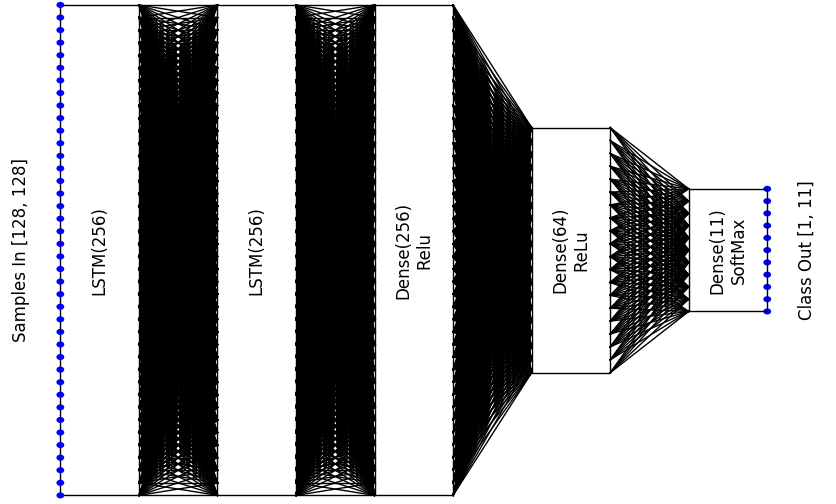}
\label{fig:lstmarch1}
\end{figure}

In our discriminative classifier we train a network to decide which traffic is being carried by the wireless network signal.  This is a K-class supervised learning problem which seeks to select which of K traffic types is currently the primary network traffic behavior in focus.  We implement both a CLDNN \cite{cldnn}, or a network formed by a sequence of convolutional layers, LSTM layers, and finally fully-connected layers, as well as a LSTM followed by fully connected layers.  The architecture for the latter is shown in \ref{fig:lstmarch1}.

Since few benchmark data sets exist in this domain, we publish our data sets on radioml.com and fully describe out approach for comparison.  We leverage an 2 layer LSTM followed by two fully-connected layers to perform class estimation using dropout of 0.5 between each layer.  

\subsubsection{Noiseless Training with Overlap}

We begin with the easiest case of dataset to ensure the learning capacity is actually present within the model we are proposing.  Here, we use the raw modulated signal, at very high signal to noise ratio (SNR), with no effects of frequency or sample timing offset introduced.  Additionally, our examples are each 128-symbol aligned which was a by-product of our initial training configuration but makes the task significantly easier for the network.  Lastly, in this training regime, we do not re-use example between training and test sets, but we do allow overlap between training and test sets.  That is, certain windows of data may be present at different offsets 

\begin{figure}[h]
\caption{Performance of classifier vs RNN sequence length}
\centering
\includegraphics[width=0.5\textwidth]{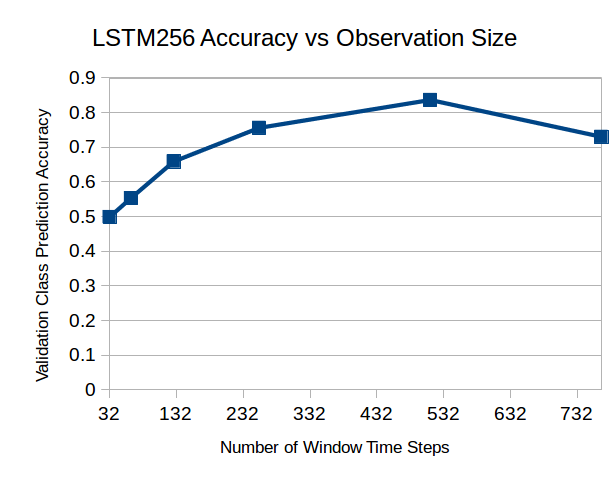}
\label{fig:perflen}
\end{figure}

In this case, we select an input tensor shape of Nx2x128 where we search over a range of N values to determine the best number of time steps for performance, shown in figure \ref{fig:perflen}

\begin{figure}[h]
\caption{Best LSTM256 confusion with RNN length of 512 time-steps }
\centering
\includegraphics[width=0.5\textwidth]{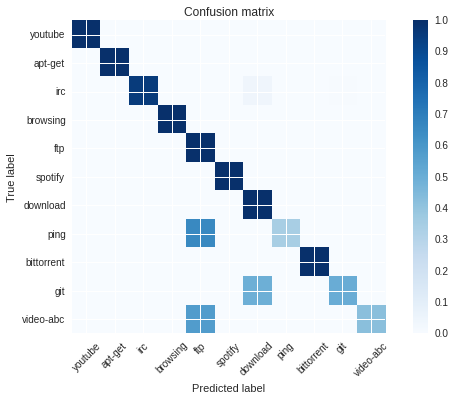}
\label{fig:bestconf}
\end{figure}

\begin{table*}
\caption{Performance measurements on varying sequence lengths} \label{tab:data}
\medskip
\centering
\begin{center}
 \begin{tabular}{||c c c c c c c||} 
 \hline
 Sequence Length & Val. Loss & Val. Accuracy & $N_{samples}$ & $N_{symbols}$ & $N_{bits}$ & Sec/Epoch \\
 \hline\hline
 32 & 1.2126 & 0.498805 & 1120 & 140  & 280 & 5 \\
 \hline
 64 &	1.0386	& 0.553546	& 2144	& 268	& 536	& 18 \\
 \hline
128	& 0.7179	& 0.65894	& 4192	& 524	& 1048	& 17 \\
\hline
256	& 0.4586	& 0.75621	& 8288	& 1036	& 2072	& 29 \\
\hline
512	& 0.2711	& 0.836535	& 16480	& 2060	& 4120	& 38 \\
\hline
768	& 0.5328	& 0.730413	& 24672	& 3084	& 6168	& 27 \\
 \hline
\end{tabular}
\end{center}
\label{tab:perftable}
\end{table*}

We find our best accuracy performance to be obtained when using 512 time-steps of 2x128 samples into the LSTM.  Details of different sequence length evaluation are detailed in table \ref{tab:perftable}. For a sequence length of 512, we obtain a confusion matrix of our best performance classification accuracy in figure \ref{fig:bestconf}, with an overall accuracy of ~84\%, with mostly-diagonal, accurate classification performance other than a few somewhat confused classes.  It is important to note that some error is inherent in the data set however as any given time window in the data may or may not have packets representing the traffic behavior of interest, we are looking at quite small windows of time here.


\subsubsection{Training with Channels and no Overlap}

To fully differentiate training and test sets, we need to fully remove overlap between example drawn from each.  In this section, we partition the original time series into hard partitions of 250,000 samples, each assigned to either training or test, and then draw examples from within these bounds for training and test.  This ensure that we are learning generalizable sequence features rather than specific window examples which may be used to recall one class.  Additionally, we consider two forms of the input signals, one we call "clean", which represents the same high-SNR signal without frequency offset or timing offset, and one which we call "channel" which applies the channel effects of additive white Gaussian noise, random frequency offset, and random timing offset.  The latter has a signal to noise ratio of around 20dB, still quite high, but reasonably realistic and much lower than clean version.  Lastly, we relax the effect of beginning on 128-symbol aligned offsets, we consider two values for "offset\_modulo": 1, where we may be begin on any offset, and 256, where we begin 128-symbol (256 -sample) aligned, to consider the additional effect of this assumption on the classification task.  These assumptions make the task significantly more difficult.

Since numerous architectures exist to evaluate on this task, and searching over them is a laborious and compute-time intensive task, we introduce a tool, still in very early form called dist\_hyperas \cite{dist_hyperas}, to help in searching for optimal hyper-parameters within an architecture over a number of different GPU instances.  

\begin{figure*}[!ht]
\caption{Trade Search 1}
\centering
\includegraphics[width=1.0\textwidth]{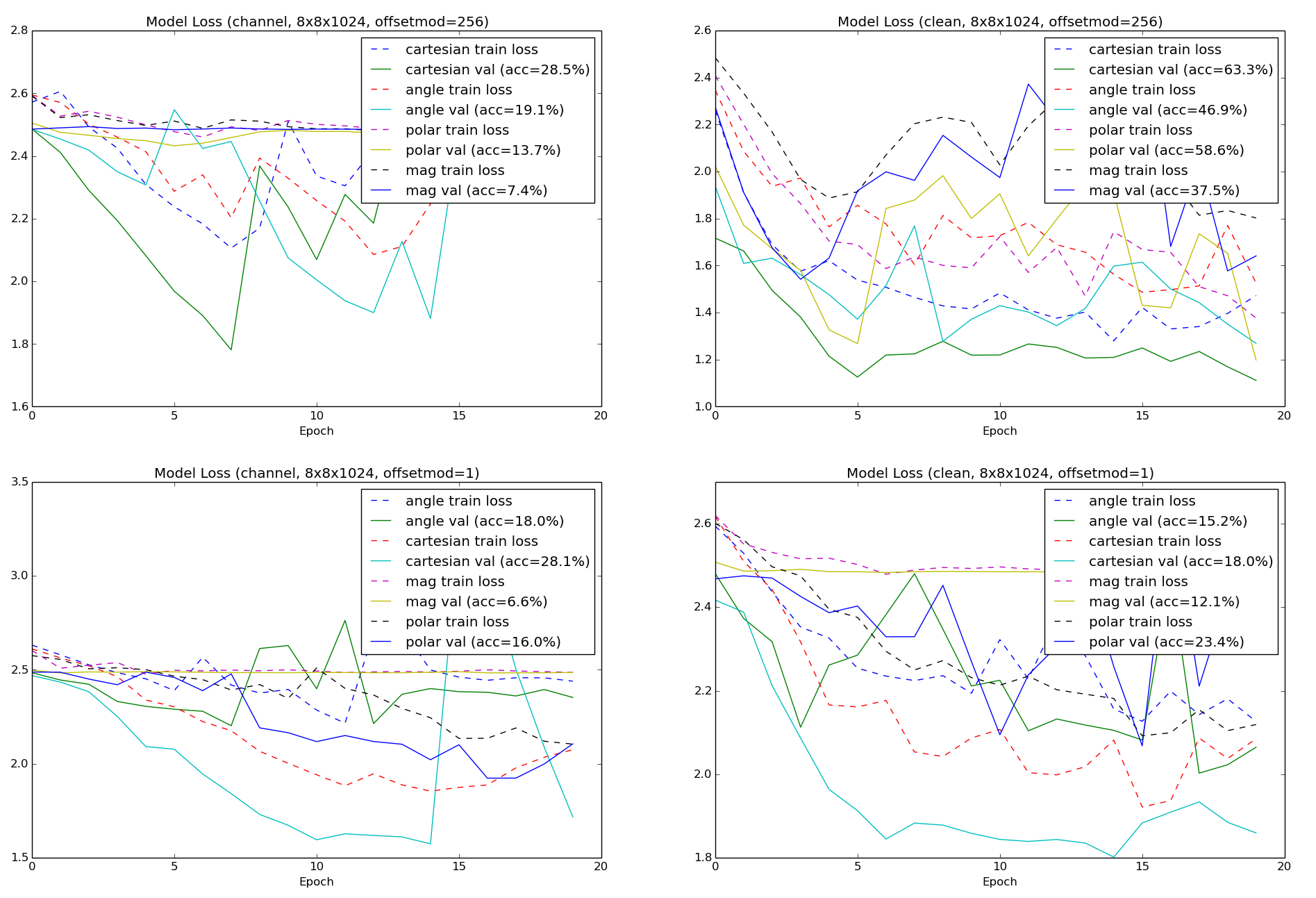}
\label{fig:trade1}
\end{figure*}

In our first trade, we evaluate the performance of input representation, channel effects and stride on our model using an 8x8x1024 input tensor shape.  The loss curves and final accuracy for each model tested is shown in figure \ref{fig:trade1}.  In this case, we obtain our best performance with a channel using the Cartesian I/Q input representation, and the offset\_modulo doesn't seem to have a huge impact when a real channel is considered.  (It has a much larger impact on performance with the clean signal).  Best performance with a channel is around 28.5\% while without a channel it is around 63.3\%.  

\begin{figure*}[!ht]
\caption{Trade Search 2}
\centering
\includegraphics[width=0.9\textwidth]{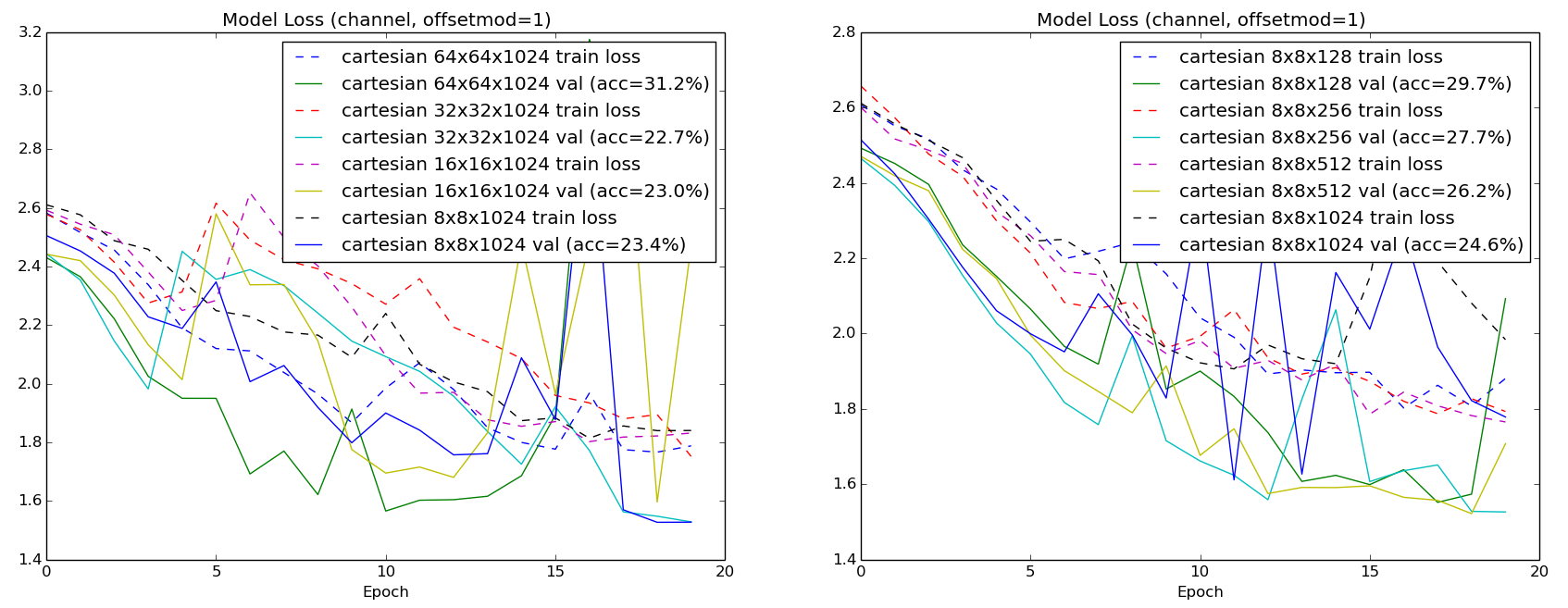}
\label{fig:trade2}
\end{figure*}

In our second trade, we consider only Cartesian I/Q inputs and an offset\_mod of 1.  In this case we trade the sequence length (number of time-steps) against the size and stride of the window used.  The results are shown in figure \ref{fig:trade2}.  In this case, we seem to obtain out best performance with a window size of L=64 and a sequence length of N=1024 giving a classification accuracy of around 31.2\% with realistic channel and sampling conditions.  We are still investigating larger models and additional hyper-parameter combinations but large LSTM architectures require large memory footprints currently, near/at the limitations of our Titan X, and training takes significant compute-time.  In the future we hope to find smarter ways to live within these limitations.

We believe some additional performance could be gained from architecture searches, but also from improved fundamental techniques described below to help cope with channel variation.

\subsection{Generative Model Training}

\begin{figure*}[!ht]
\caption{Best LSTM256 prediction of IRC sequence }
\centering
\includegraphics[width=0.9\textwidth]{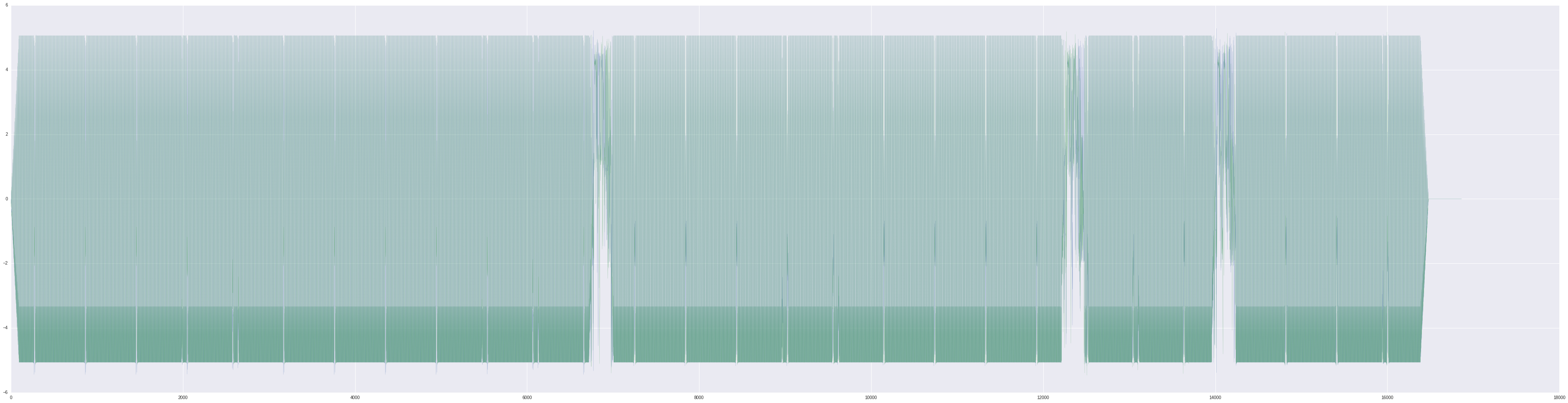}
\label{fig:seqpred}
\end{figure*}

\begin{figure*}[!ht]
\caption{Best LSTM256 prediction of Spotify sequence }
\centering
\includegraphics[width=1.0\textwidth]{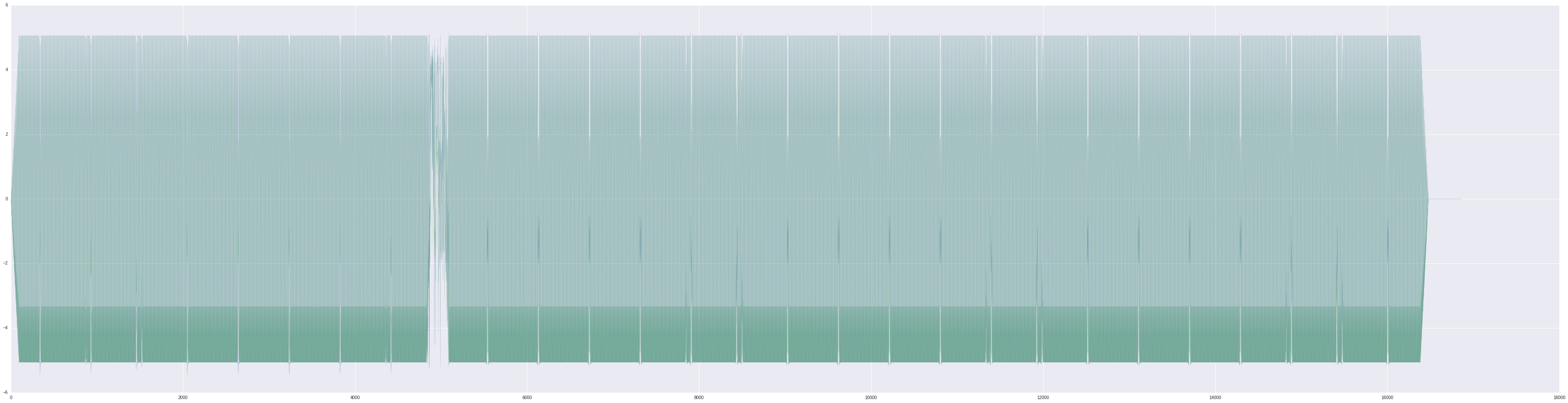}
\label{fig:seqpred2}
\end{figure*}

\begin{figure}[!ht]
\caption{Best LSTM256 generative regression network}
\centering
\includegraphics[width=0.5\textwidth]{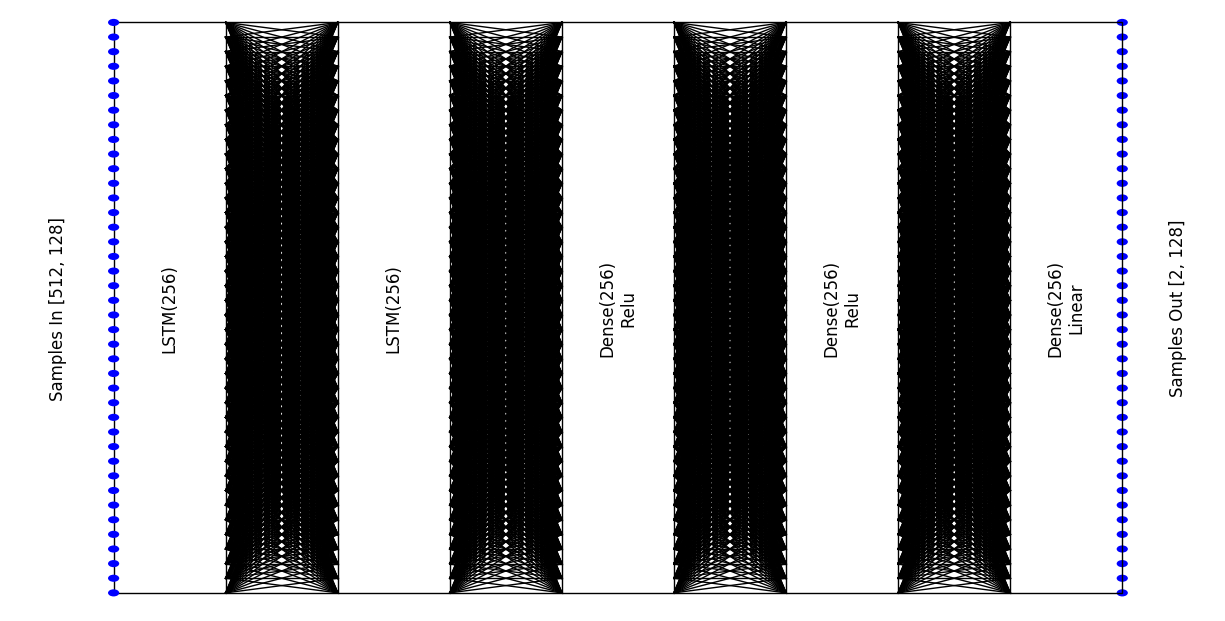}
\label{fig:net2}
\end{figure}

We employ a simple first order generative model shown in figure \ref{fig:net2} which predicts the next time-step window given N previous time-step windows as a regression task.  We train network parameters using mean squared error (MSE) of real output sample values with a linear output layer activation function.

In figure \ref{fig:seqpred} we show a modulated radio data signal where the first half is ground truth from an IRC sequence example and the second half of samples is predicted from a generative model using the recurrent neural network model described herein. 

Visually comparing the predicted samples to those from the baseline example, we can see it correctly predicts the HDLC idle pattern, the equal-width framing pattern, and some semblance of data bursts occurring within the generative sample data region.  This is somewhat impressive for a completely naive model training effort based only on a handful of available training data sequences, considering it has no expert knowledge of the modulation, preamble structure, the HDLC protocol, or the application on top.

In figure \ref{fig:seqpred2} we show a similar sequence for Spotify music streaming where the first half is a real example and the second half is generated from the model.  In this case we can see the generated HDLC idle pattern, equi-distance frame preamble, but no additional data bursts occurring.

In future work we plan to use a Generative Adversarial Network (GAN) architecture \cite{gan} approach to improving our generative model realism by introducing a critic/discriminator model.  This technique has proven extremely effective in the image domain by introducing a feedback loop of real/generated discriminator critique against the generator output to form a reinforcing learning process by which both models improve each other an result in more realistic generative outputs.  

Two extremely promising recent approaches to time-series generation we believe are extremely applicable here for future work are presented in \cite{seqgan} and \cite{wavenet}.

\section{Conclusion}

We have shown in this work that recurrent neural network models can be readily used in high level radio protocol sequence recognition from pre-demodulated radio signal data for both discriminative labeling and generative emulation tasks.  

We have demonstrated baseline performance for both tasks which works quite well under ideal conditions (high SNR, no frequency of sample rate offset).  However, introducing realistic channel effects makes the task significantly more difficult and significantly reduces model performance.

The channel variations to the sequence introduced over a wireless channel in sample rate offset, frequency offset, and channel delay spread make learning sequence models from raw data difficult, but a number of ideas exist which may help alleviate this problem such as allowing attention models to cononicalize the channel effects out \cite{rtn} and the introduction of heavy channel regularization during training as described in \cite{chanae}.

These results have significant impact into sequence and protocol recognition learning for numerous cognitive and traditional radio applications.  By providing a robust method for protocol identification learning which is data and experience driven, numerous future radio allocation, QoS, scheduling and decision making algorithms can make intelligent decisions about how to prioritize and allocate radio data within a larger resource constrained multi-user cognitive radio networked system.  

\section*{Acknowledgment}

The authors would like to thank the Bradley Department of Electrical and Computer Engineering at the Virginia Polytechnic Institute and State University, the Hume Center, and DARPA all for their generous support in this work.

This research was developed with funding from the Defense Advanced Research Projects Agency's (DARPA) MTO Office under grant HR0011-16-1-0002. The views, opinions, and/or findings expressed are those of the author and should not be interpreted as representing the official views or policies of the Department of Defense or the U.S. Government.

\printbibliography

\end{document}